\documentclass[10pt,twocolumn,letterpaper]{article}

\usepackage{iccv}
\usepackage{times}
\usepackage{epsfig}
\usepackage{graphicx}
\usepackage{amsmath}
\usepackage{amsthm}
\usepackage{amssymb}
\usepackage{multirow}
\usepackage{booktabs}
\usepackage{bbm}

\usepackage[pagebackref=true,breaklinks=true,letterpaper=true,colorlinks,bookmarks=false]{hyperref}

\newtheorem{definition}{Definition}

\newcommand{\blue}[1]{\textcolor{blue}{#1}}

\usepackage{xcolor}
\definecolor{lightblue}{RGB}{73, 116, 150}
\definecolor{codekey}{RGB}{255,140,0}
\definecolor{deepblue}{RGB}{25, 60, 100}

\iccvfinalcopy 

\begin{document}

\title{Generalized Categories Discovery for Long-tailed Recognition}

\author{Ziyun Li\\
Hasso Plattner Institut\\
University of Potsdam\\
{\tt\small ziyun.li@hpi.de}
\and
Christoph Meinel\\
Hasso Plattner Institut\\
University of Potsdam\\
{\tt\small christoph.meinel@hpi.de}
\and
Haojin Yang\\
Hasso Plattner Institut\\
University of Potsdam\\
{\tt\small haojin.yang@hpi.de}
}

\maketitle
\ificcvfinal\thispagestyle{empty}\fi

\begin{abstract}

Generalized Class Discovery (GCD) plays a pivotal role in discerning both known and unknown categories from unlabeled datasets by harnessing the insights derived from a labeled set comprising recognized classes. 
A significant limitation in prevailing GCD methods is their presumption of an equitably distributed category occurrence in unlabeled data. Contrary to this assumption, visual classes in natural environments typically exhibit a long-tailed distribution, with known or prevalent categories surfacing more frequently than their rarer counterparts. 
Our research endeavors to bridge this disconnect by focusing on the long-tailed Generalized Category Discovery (Long-tailed GCD) paradigm, which echoes the innate imbalances of real-world unlabeled datasets. 
In response to the unique challenges posed by Long-tailed GCD, we present a robust methodology anchored in two strategic regularizations: (i) a reweighting mechanism that bolsters the prominence of less-represented, tail-end categories, and (ii) a class prior constraint that aligns with the anticipated class distribution. 
Comprehensive experiments reveal that our proposed method surpasses previous state-of-the-art GCD methods by achieving an improvement of approximately \textbf{6 - 9\%} on ImageNet100 and competitive performance on CIFAR100. 
\end{abstract}


\section{Introduction}
In the context of modern machine learning approaches, the effectiveness of models is heavily dependent on their training data, especially when trained on large datasets that have been annotated by humans. This limitation restricts their effectiveness in real-world situations where encountering unannotated data from previously unseen categories is common. 
To fill this research gap, scholars have explored several avenues. One approach is open-set recognition \cite{scheirer2012toward}, which aims to determine if an unlabeled piece of data belongs to a class we've seen before. Another is the Novel Category Discovery (NCD) \cite{han2019learning,han2021autonovel,fini2021unified,li2022closer}. However, early NCD methods were flawed, assuming all unlabelled images represented entirely new categories. This isn't typically the case in real-world scenarios. Recognizing this limitation, the Generalized Category Discovery (GCD) \cite{vaze2022generalized} framework was developed. GCD understands that unlabeled data can be from both previously known or entirely new categories, making it a more realistic approach for diverse data settings.

A pivotal challenge in GCD is identified: the \textit{lack of information about occurrence frequency} of categories, as mention in \cite{li2023imba}, whether known or unknown, forming the crux of our research. Notably, much of the existing research, as seen in \cite{vaze2022generalized, sunopencon, cao2022open, rizve2022openldn}, neglects this frequency issue. 
Long-tailed GCD aligns more accurately with the real-world, emphasizing the long-tailed distribution of visual classes with known classes like  dominate the distribution's ``head", whereas unknown classes being rarer, populate the sparse ``tail".
We present a novel method to address the challenges inherent to long-tailed GCD. One predominant challenge in long-tailed GCD is the tendency of the system to exhibit bias towards head classes, often at the expense of tail/unknown classes. This is manifested when tail or unknown samples are frequently misclassified as head or known classes. The primary reason for this is the model's inclination to learn predominantly from the patterns of the more prevalent classes, which in turn compromises the accuracy for the minority classes.

To rectify this, our approach diverges from the ImbaGCD \cite{li2023imba} of utilizing optimal transport alignment in accordance with the estimated prior distribution. 
Instead, we introduce two strategic constraints:
(i) Reweighting of tail/unknown classes: We amplify the weight attributed to the tail distribution.
(ii) Alignment of the estimated prior distribution: We impose constraints on the estimated class distribution.
Compared to ImbaGCD \cite{li2023imba}, our method substantially reduces computational complexity by eliminating the EM and Sinkhorn iterations, which are used to solve the optimal transport function. Concurrently, we observe a performance improvement of 6 - 9\% on ImageNet100 and achieve competitive results on CIFAR100.

\section{Problem Setup}
%

\begin{definition}[Long-tailed Generalized Category Discovery]
Let $\mathbb{P}_{\mathbf{X}_l,Y_l}$ and $\mathbb{Q}_{\mathbf{X}_u,Y_u}$ represent labeled and unlabeled probability measures on $\mathcal{X}_l \times \mathcal{Y}_l$ and $\mathcal{X}_u \times \mathcal{Y}_u$ respectively, with $\mathcal{Y}_l\subset\mathcal{Y}_u$ ($\mathcal{Y}_u$ represents all classes in the unlabeled set, comprising known classes $\mathcal{Y}_l$ and unknown classes $\mathcal{Y}_n=\mathcal{Y}_u\backslash\mathcal{Y}_l$) and $\mathcal{Y}_n=\mathcal{Y}_u\backslash\mathcal{Y}_l$. Given datasets $\mathcal{L}_n$ and $\mathcal{U}_m$ sampled from these measures, the goal of GCD is to predict labels for $\mathcal{U}_m$, which can belong to known or unknown classes.
In long-tailed GCD, the class distributions in the unlabeled set are skewed such that \(\mathbb{P}_{Y_{{l}}}(i) > \mathbb{P}_{Y_{{n}}}(j)\) for every \(i \in \mathcal{Y}_l\) and \(j \in \mathcal{Y}_n\).
\end{definition}  

\section{Method}


\subsection{Class Prior Estimation}
\paragraph{Moving-average distribution update} 
We suggest employing model predictions for class prior estimation following \cite{guo2017calibration,li2023imba}. However, due to potential inaccuracies and biases in early training stages, we propose a moving-average update mechanism to enhance reliability.
Starting with a uniform class prior $\mathbf{r}=[1 / C, \ldots, 1 / C]$, we iteratively refine the distribution per epoch.
\begin{gather*}
   \small
     \mathbf{r} := \mu \mathbf{r}+(1-\mu) \mathbf{z}, \\
     \quad \text { where } \mathbf{z}_{j}=\frac{1}{n} \sum_{i=1}^{n} \mathbbm{1}\left(j=\arg \max _{j^{\prime}} f_{j^{\prime}}\left(\mathbf{x}_{i}\right)\right), \mu \in[0,1]
  \end{gather*}

The class prior is continuously updated through a linear function, resulting in more stable training dynamics. As the training progresses, the model's accuracy improves, making the estimated distribution increasingly dependable.

\subsection{Representation Improvement}
\label{subsec:representation}
To enhance model representation, we adopt unsupervised contrastive learning \cite{chen2020simple, he2020momentum} for unlabeled data and supervised contrastive learning \cite{khosla2020supervised} for labeled data, as in \cite{sunopencon,cao2022open}. Specifically, let $\mathbf{v}_{i}$ and $\mathbf{v}_{i}^{\prime}$ represent features from two views (random augmentations) of the same image within a mini-batch $B$. The instance-level unsupervised contrastive loss and the supervised constrastive loss are defined as follows:
\begin{equation}
\small
   \begin{aligned}
      \mathcal{L}^{(i)}_{ins}=\frac{1}{|B_u|}\sum_{i \in B_u}-\log \frac{\exp \left(\mathbf{v}_{i} \cdot \mathbf{v}_{i}^{\prime} / \tau\right)}{\sum_{i}^{i \neq j} \exp \left(\mathbf{v}_{i} \cdot \mathbf{v}_{j} / \tau\right)},
   \end{aligned}
\end{equation}
\begin{equation}
   \small
   \mathcal{L}^{(i)}_{sup}=\frac{1}{\left|B_{l}\right|}\sum_{i \in B_{l}}\frac{1}{|\mathcal{N}^{(i)|}} \sum_{q \in \mathcal{N}^{(i)}} -\log \frac{\exp \left(\mathbf{v}_{i} \cdot \mathbf{v}_{q} / \tau\right)}{\sum_{i}^{i \neq j}\exp \left(\mathbf{v}_{i} \cdot \mathbf{v}_{j} / \tau\right)},
   \end{equation}
where $B_u$ and $B_l$ represent the unlabeled and labeled subsets of mini-batch $B$, respectively. Furthermore, $\mathcal{N}^{(i)}$ refers to the indices of other images in the batch sharing the same label, and $\tau$ signifies a temperature parameter.

\subsection{Constraints on Class Distribution}
Unlike ImbaGCD \cite{li2023imba} involve sophisticated techniques such as optimal transport for aligning class prior distribution, the EM algorithm for generating pseudo labels, and updating prototypes.
In contrast, our approach is geared towards simplicity and directness, focusing on addressing the inherent challenges of class imbalances, especially in the context of tail classes. 

We propose two constraints:
(i) \textit{Reweighting tail samples}:
A prevalent challenge in dealing with imbalanced datasets is the difficulty of learning patterns from tail classes, given their notably fewer samples compared to the head classes. 
One direct and effective way to mitigate this is through sample reweighting. 
Previous works have demonstrated the efficacy of reweighting schemes in various contexts \cite{cao2019learning, park2021influence,li2021autobalance}. 
To effectively balance the information derived from tail classes with the rest, we incorporate a constraint between model predictions $\mathbf{q}$ and 
the uniform distribution $\mathbf{u}$:
\begin{equation}
   \mathrm{H}(\mathbf{q},\mathbf{u}) = \mathbf{q} \log \mathbf{u}, \quad \text {where } \mathbf{u} = [1/C,...,1/C]
\end{equation}
(ii) \textit{Aligning Class Prior:}
While optimal transport methods like Sinkhorn have their merits, they can also be reduced to simpler forms such as KL divergence \cite{peyre2017computational}. For our work, in pursuit of simplicity without significant trade-off in performance, we employ entropic regularization between the estimated distribution and the output. It is represented as:
\begin{equation}
\mathrm{H}(\mathbf{q},\mathbf{r}) = \mathbf{q} \log \mathbf{r}
\end{equation}



\subsection{Overall Loss Objective}
\label{subsec:overall_loss}
The overall loss objective is a weighted sum of the unsupervised and supervised contrastive losses and two regularizations:
\begin{equation}
   \label{eq:overall}
   \small
\mathcal{L_\text{overall}}= \mathcal{L}_{ins}+\lambda\mathcal{L}_{sup} +  \alpha\mathrm{H}(\mathbf{q},\mathbf{r}) + \beta\mathrm{H}(\mathbf{q},\mathbf{u})
\end{equation}

\begin{table*}[t]
   \centering
   \setlength{\tabcolsep}{12pt}
   \caption{Performance comparison of various methods on ImageNet100 under different imbalanced factors $\rho$, ($\rho=0.5$ is original GCD setting). 
   Our method consistently outperforms others on novel class under imbalanced settings and achieves competitive performance on the balanced setting.
  We report the mean and standard deviation of the
  clustering accuracy across 3 runs for multiple methods. 
  The higher mean
  value is presented in bold, while the results within standard deviation of the average accuracy are not bolded.}
   \label{tab:ImageNet100}
   \begin{tabular}{@{}lccllll@{}}
      \toprule
   \multirow{2}{*}{IMF} & \multirow{2}{*}{Metrics} & \multicolumn{4}{c}{Methods}                     \\ \cmidrule(l){3-7} 
                                       &                          & GCD                  & \multicolumn{1}{c}{ORCA} & \multicolumn{1}{c}{OpenCon} & \multicolumn{1}{c}{ImbaGCD} &  {Ours}\\
                                       \midrule
   \multirow{4}{*}{$\rho=0.5$}
   & All                   & 77.12\tiny{$\pm$0.56} & 74.93\tiny{$\pm$0.34} & 82.22\tiny{$\pm$0.56} & 81.90\tiny{$\pm$0.67} & 81.06\tiny{$\pm$0.39}\\ 
   & Known                  &  87.02\tiny{$\pm$0.36} & 89.21\tiny{$\pm$0.06} & 90.65\tiny{$\pm$0.04} & {91.19}\tiny{$\pm$0.15} & 91.03\tiny{$\pm$0.03}\\
   & Un1              & 57.48\tiny{$\pm$0.76} & 67.18\tiny{$\pm$0.27} & 78.12\tiny{$\pm$0.80} & 77.89\tiny{$\pm$0.88} & 77.22\tiny{$\pm$0.25}\\
   & Un2              &  42.68\tiny{$\pm$0.77} & 65.43\tiny{$\pm$0.37} & 78.01\tiny{$\pm$0.83} & 77.79\tiny{$\pm$0.97} & 77.68\tiny{$\pm$0.30}\\
                                 \midrule
                                    \multirow{4}{*}{$\rho=1$}   
   & All                   &  63.06\tiny{$\pm$0.66} & 68.01\tiny{$\pm$0.31}  & 82.44\tiny{$\pm$0.24} &  82.34\tiny{$\pm$0.39} & 82.28\tiny{$\pm$0.68}\\					
   &	Known                 &  88.80\tiny{$\pm$0.43} & 88.99\tiny{$\pm$0.05}  & 90.62\tiny{$\pm$0.12} &  90.56\tiny{$\pm$0.18} & \textbf{91.30}\tiny{$\pm$0.10}\\					
   &	Un1         &  37.29\tiny{$\pm$1.11} & 47.68\tiny{$\pm$0.70}  & 74.45\tiny{$\pm$0.36} &  74.56\tiny{$\pm$0.79} &75.06\tiny{$\pm$1.00}\\					
   &	Un2             &  31.16\tiny{$\pm$1.48}  & 47.28\tiny{$\pm$0.66}  & 74.35\tiny{$\pm$0.38} &   74.24\tiny{$\pm$0.89} &73.39\tiny{$\pm$1.38}\\
   \midrule
   \multirow{4}{*}{$\rho=5$}  
   & All      &  77.29\tiny{$\pm$0.17}  & 76.79\tiny{$\pm$0.13} & 81.87\tiny{$\pm$0.23} & {83.01}\tiny{$\pm$0.11} &\textbf{86.14}\tiny{$\pm$0.08} \blue{\small{($+$ 3.13)}}\\ 
   & Known     &  87.51\tiny{$\pm$0.21}  & 89.02\tiny{$\pm$0.12} &  {90.77}\tiny{$\pm$0.10} & 88.89\tiny{$\pm$0.06} &\textbf{91.69}\tiny{$\pm$0.31} \blue{\small{($+$ 2.80)}}\\
   & Un1&  23.83\tiny{$\pm$0.24}  & 20.19\tiny{$\pm$0.55} & 39.76\tiny{$\pm$1.4} & {54.35}\tiny{$\pm$0.38} &\textbf{61.17}\tiny{$\pm$0.91} \blue{\small{($+$ 6.82)}}\\
   & Un2 & 14.80\tiny{$\pm$0.23}  & 16.42\tiny{$\pm$0.73} & 37.93\tiny{$\pm$1.3} & {53.97}\tiny{$\pm$0.38} &\textbf{59.10}\tiny{$\pm$0.59} \blue{\small{($+$ 4.21)}}\\
   \midrule
   \multirow{4}{*}{$\rho=10$}  
   & All      & 83.32\tiny{$\pm$0.15}  & 81.98\tiny{$\pm$0.03} & {84.60}\tiny{$\pm$0.10} & 83.23\tiny{$\pm$0.17} & \textbf{87.44}\tiny{$\pm$0.31} \blue{\small{($+$ 4.21)}}\\
   & Known     &  89.71\tiny{$\pm$0.03}  & 89.24\tiny{$\pm$0.02} & {90.85}\tiny{$\pm$0.12} & 87.48\tiny{$\pm$0.23} &\textbf{91.51}\tiny{$\pm$0.11} \blue{\small{($+$ 4.03)}}\\
   & Un1& 22.17\tiny{$\pm$0.18}  & 16.94\tiny{$\pm$0.07} & 26.70\tiny{$\pm$0.11} & {42.62}\tiny{$\pm$0.50} &\textbf{51.18}\tiny{$\pm$2.53} \blue{\small{($+$ 8.56)}}\\
   & Un2 &  11.47\tiny{$\pm$0.28}  & 10.31\tiny{$\pm$0.22} & 22.89\tiny{$\pm$0.27} & {41.23}\tiny{$\pm$0.58} &\textbf{47.61}\tiny{$\pm$2.59} \blue{\small{($+$ 6.58)}}\\

   \bottomrule               
   \end{tabular}
   \end{table*}

\section{Experiments}
In this section, we conduct an experimental analysis of our proposed method under both original and varying imbalanced scenarios.
\subsection{Setup}
\paragraph{Datasets} 

Experiments were conducted on CIFAR10, CIFAR100, and ImageNet100. Classes were split evenly into known and unknown. From the known classes, 50\% were randomly selected as a balanced labeled dataset. In the unlabeled set, the class sample sizes were imbalanced, adjusted by factor \(\rho\), defined as the ratio \(\rho=\frac{n_{k}}{n_{u}}\) where \(n_{k}\) and \(n_{u}\) are the sample sizes of known and unknown classes. We considered \(\rho \in\{0.5, 1, 5, 10\}\). Previous studies typically had a fixed imbalance of \(\rho=0.5\) by using the remaining 50\% of known class samples as the unlabeled set.

\paragraph{Evaluation metrics} 

We use the evaluation method from \cite{cao2022open,sunopencon,vaze2022generalized}, focusing on (1) overall accuracy, (2) accuracy for known classes, and, as in ImbaGCD \cite{li2023imba}, on novel data under settings: (3) \textbf{unknown-aware} and (4) \textbf{unknown-agnostic}. 
In the \textbf{unknown-aware} evaluation, we separate unlabeled samples from unknown classes and cluster them. However, this might not mirror real-life situations where telling apart known and unknown classes isn't straightforward. Hence, we introduce the \textbf{unknown-agnostic} metric that evaluates without pre-classifying samples, offering a more genuine and neutral assessment.

\paragraph{Implementation details}
We use ResNet18 for CIFAR10/100 and ResNet50 for ImageNet100. Training lasts 200 epochs for CIFAR10/100 and 120 for ImageNet100, with batch sizes of 512. We apply stochastic gradient descent (momentum 0.9, weight decay \(10e-4\)), starting with a learning rate of 0.02 that decays 10x at 50\% and 75\% milestones. The momentum, \(\mu\), for class prior remains at 0.99.

\begin{table*}[t]
   \centering
   \tiny
   \setlength{\tabcolsep}{12pt}
   \fontsize{10pt}{12pt}\selectfont
   \caption{Performance comparison on CIFAR100 with varying imbalance factors $\rho$ ($\rho=0.5$ for original GCD setting). 
   Un1 represents unknown-aware accuracies, while Un2 refers to unknown-agnostic accuracies.
   Our method consistently excels in novel class performance under imbalanced settings and remains competitive in balanced settings. Mean and standard deviation of clustering accuracy are reported over 3 runs for multiple methods. Higher means are bolded, while results within one standard deviation are not.}
   \label{tab:sota}
   \vspace{0.3cm}
   \begin{tabular}{@{}lcccccc@{}}
      \toprule
   \multirow{4}{*}{IMF} & \multirow{4}{*}{Metrics} & \multicolumn{5}{c}{CIFAR100} \\ \cmidrule(l){3-7}
   & & GCD         & \multicolumn{1}{c}{ORCA} & \multicolumn{1}{c}{OpenCon} & \multicolumn{1}{c}{ImbaGCD} & \multicolumn{1}{c}{Ours} \\
                                       \midrule
   \multirow{4}{*}{$\rho=0.5$}
   & All                   & 45.41\tiny{$\pm$0.13} & 55.68\tiny{$\pm$0.33}& 51.85\tiny{$\pm$0.63} & 53.51\tiny{$\pm$0.26} & 52.20\tiny{$\pm$0.30}\\ 
   & Known                  & 67.61\tiny{$\pm$0.12} & 66.41\tiny{$\pm$0.31} & 69.07\tiny{$\pm$0.29} & 68.09\tiny{$\pm$0.13} & 70.45\tiny{$\pm$0.31}\\
   & Un1              & 34.31\tiny{$\pm$0.22} & 42.63\tiny{$\pm$0.67} & 45.76\tiny{$\pm$0.32} & \textbf{47.92}\tiny{$\pm$0.33} & 46.96\tiny{$\pm$0.35}\\
   & Un2              & 18.12\tiny{$\pm$0.34} & 38.95\tiny{$\pm$0.76} & 42.11\tiny{$\pm$0.54} & {46.22}\tiny{$\pm$0.33} &46.25\tiny{$\pm$0.25} \\
                                 \midrule
   \multirow{4}{*}{$\rho=1$}   
   & All                   & 48.36\tiny{$\pm$0.08} & 47.37\tiny{$\pm$0.52} & 53.20\tiny{$\pm$0.33} & 54.06\tiny{$\pm$0.45}& 55.82\tiny{$\pm$0.40}\\					
   &	Known                 & 71.48\tiny{$\pm$0.24} & 65.14\tiny{$\pm$0.17} & 68.00\tiny{$\pm$0.06} & 67.98\tiny{$\pm$0.37} & 68.00\tiny{$\pm$0.35}\\					
   &	Un1         & 25.24\tiny{$\pm$0.07} & 34.93\tiny{$\pm$1.04} & 43.58\tiny{$\pm$0.34} & 43.39\tiny{$\pm$0.59} & \textbf{44.54}\tiny{$\pm$0.40}\\					
   &	Un2             & 12.02\tiny{$\pm$0.46} & 29.61\tiny{$\pm$1.02} & 38.40\tiny{$\pm$0.65} & {38.76}\tiny{$\pm$0.9} & \textbf{42.65}\tiny{$\pm$0.89}\\			
                                 \midrule
   \multirow{4}{*}{$\rho=5$}  
   & All      & 63.13\tiny{$\pm$0.12} & 56.37\tiny{$\pm$0.12} & 61.84\tiny{$\pm$0.29} & 59.48\tiny{$\pm$0.48} & \textbf{62.85}\tiny{$\pm$0.53}\\ 
   & Known     & 70.96\tiny{$\pm$0.13} & 64.40\tiny{$\pm$0.11} & 69.37\tiny{$\pm$0.23} & 67.82\tiny{$\pm$0.06}& \textbf{70.15}\tiny{$\pm$0.05}\\
   & Un1& 24.04\tiny{$\pm$0.12} & 25.36\tiny{$\pm$0.40} & 35.06\tiny{$\pm$0.34} & {37.87}\tiny{$\pm$1.59} & 36.72\tiny{$\pm$1.23}\\
   & Un2 & 7.18\tiny{$\pm$0.34}  & 16.18\tiny{$\pm$0.23} & 24.21\tiny{$\pm$0.67} & \textbf{27.64}\tiny{$\pm$2.26} &26.36\tiny{$\pm$1.98} \\
   \midrule
   \multirow{4}{*}{$\rho=10$}  
   & All      & 66.21\tiny{$\pm$0.23} & 59.61\tiny{$\pm$0.27} & 65.05\tiny{$\pm$0.29} & 63.21\tiny{$\pm$0.05} & 62.84\tiny{$\pm$0.06}\\ 
   & Known     & 70.36\tiny{$\pm$0.31} & 64.27\tiny{$\pm$0.32} & 69.80\tiny{$\pm$0.23} & 67.82\tiny{$\pm$0.07}& 66.78\tiny{$\pm$0.04}\\
   & Un1& 24.74\tiny{$\pm$0.45} & 26.21\tiny{$\pm$0.69} & 33.01\tiny{$\pm$0.34} & {34.87}\tiny{$\pm$0.70} & 34.08\tiny{$\pm$0.54}\\
   & Un2 & 7.28\tiny{$\pm$0.21}  & 13.01\tiny{$\pm$0.34} & 17.57\tiny{$\pm$0.67} & {21.68}\tiny{$\pm$0.29} & 23.52\tiny{$\pm$0.32}\\
                                 \bottomrule                            
   \end{tabular}
   \label{tab:main_experiment}
\end{table*}

\subsection{Compared with SOTA}
\paragraph{Baselines} 

We benchmark our method against four leading GCD techniques:
(1) GCD \cite{vaze2022generalized}, a foundational GCD framework that clusters unknown categories without exhaustive labeling.
(2) ORCA \cite{cao2022open}, melds supervised and unsupervised learning, addressing open-world class imbalances.
(3) OpenCon \cite{sunopencon}, uses contrastive learning to adapt to new categories with minimal supervision.
(4) ImbaGCD \cite{li2023imba}, employs the EM algorithm, combining Sinkhorn-based pseudo-labeling with contrastive updates. The class prior and prototype are refreshed every epoch.
Our evaluation spans original GCD, balanced, and various imbalanced scenarios for a thorough comparison.

\paragraph{Comparisons with the state-of-the-art}

In Table \ref{tab:sota}, our method stands out on ImageNet datasets, especially amidst greater imbalance. We best the leading baseline by approximately 6 - 7\% and 7 - 9\% in the unknown-aware and unknown-agnostic scenarios for $\rho=5$ and $\rho=10$, respectively. 
For CIFAR100, it also achieves competitive results across a range of imbalanced settings.
Notably, the unknown-agnostic assessment is tougher than its unknown-aware counterpart, especially in highly skewed scenarios. In these demanding conditions, our method shines brighter, with its gains in unknown-agnostic results being especially pronounced compared to the unknown-aware ones.

\begin{figure}[]
   \vspace{-0.0cm}
   \centering
   \includegraphics[width=0.8\linewidth]{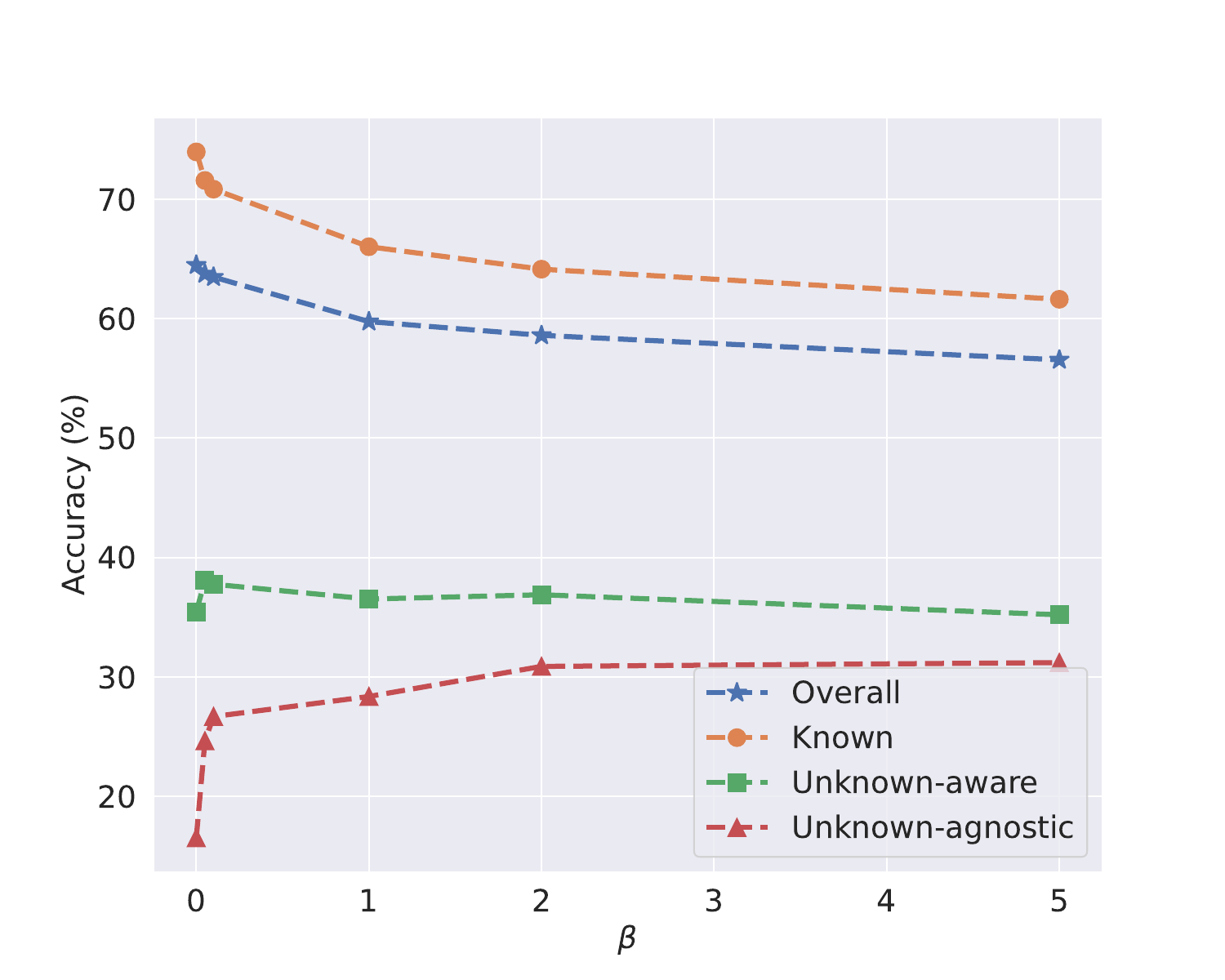}
   \caption{Performance trends of Overall, Known, Unknown-Aware, and Unknown-Agnostic accuracies across different $\beta$ values with $\alpha=0$ constant on CIFAR100 with $\rho=5$.}
   \label{fig:cifar100_beta}
   \vspace{-2mm}
\end{figure}

\begin{figure}[]
   \vspace{-0.0cm}
   \centering
   \includegraphics[width=0.8\linewidth]{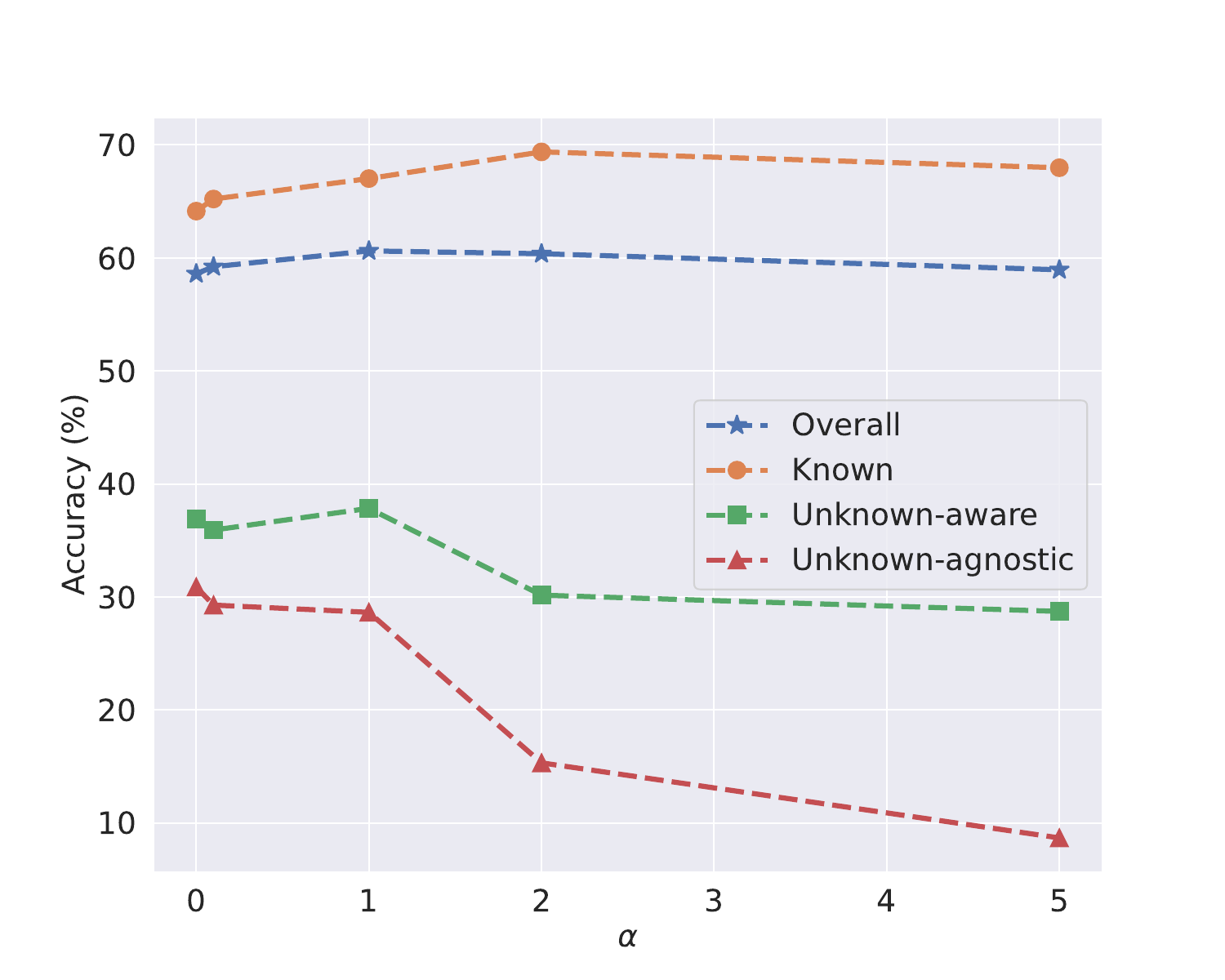}
   \caption{Performance trends of Overall, Known, Unknown-Aware, and Unknown-Agnostic accuracies across different $\alpha$ values with $\beta=2$ constant on CIFAR100 with $\rho=5$.}
   \label{fig:cifar100_alpha_2}
   \vspace{-2mm}
\end{figure}

\begin{figure}[]
   \vspace{-0.0cm}
   \centering
   \includegraphics[width=0.8\linewidth]{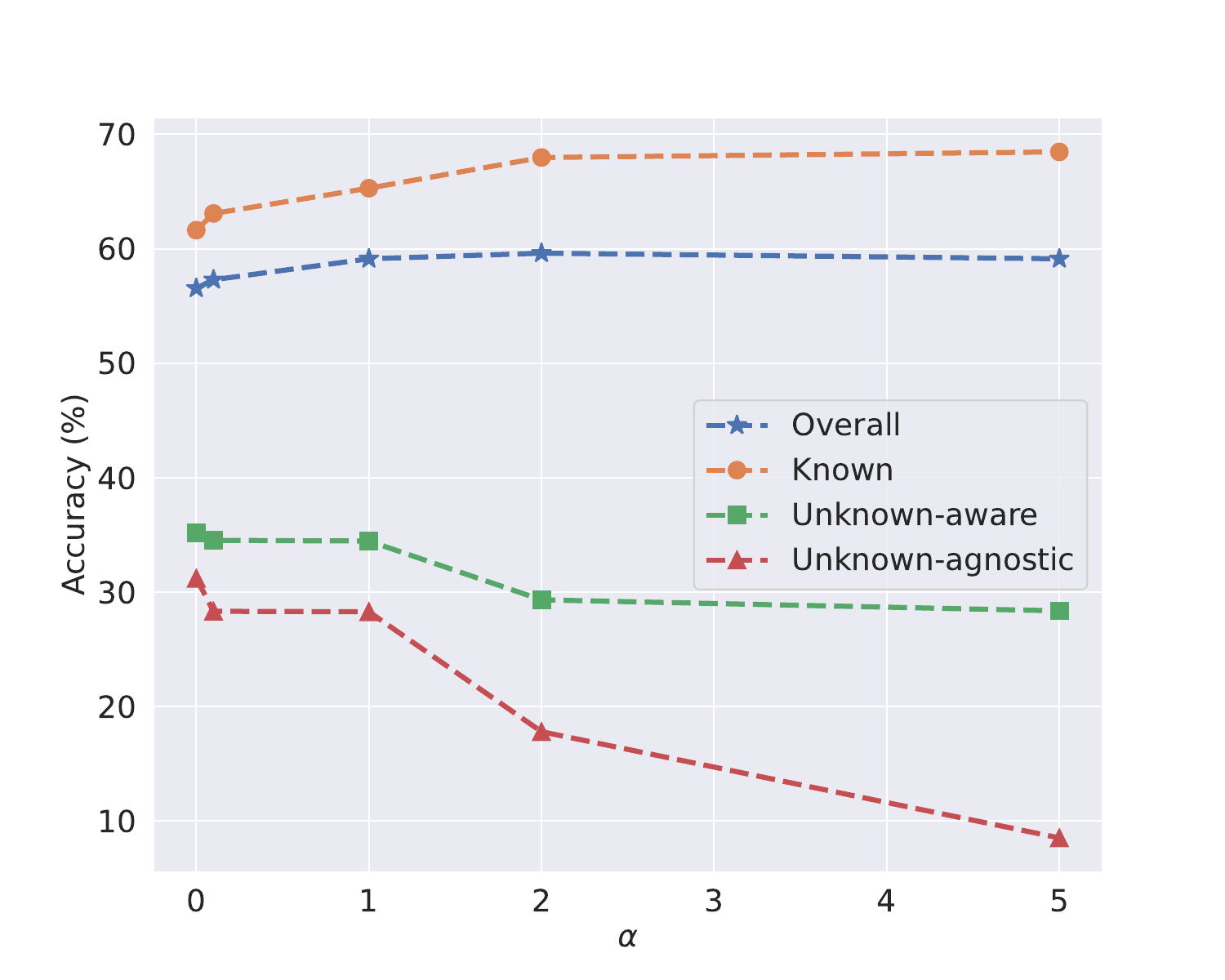}
   \caption{Performance trends of Overall, Known, Unknown-Aware, and Unknown-Agnostic accuracies across different $\alpha$ values with $\beta=5$ constant on CIFAR100 with $\rho=5$.}
   \label{fig:cifar100_alpha_5}
   \vspace{-2mm}
\end{figure}

\subsection{ Hyper-parameters Analysis}
\paragraph{Analysis of $\beta$}
We carried out experiments on the CIFAR100 dataset with an imbalance factor of $\rho=5$. By fixing $\alpha=0$, we analyzed the impact of varying $\beta$ on different accuracy metrics as shown in Figure \ref{fig:cifar100_beta}. Our results show that a higher $\beta$ enhances the unknown-aware and unknown-agnostic accuracies. However, this increase comes at the expense of known accuracy.

\paragraph{Analysis of $\alpha$}
To understand the influence of $\alpha$, we set $\beta=2$ (Figure \ref{fig:cifar100_alpha_2}) and $\beta=5$ (Figure \ref{fig:cifar100_alpha_5}), respectively, and observed how changes in $\alpha$ affected the accuracy metrics. In contrast to the behavior of $\beta$, higher values of $\alpha$ led to an improvement in known accuracy but resulted in reductions in both unknown-aware and unknown-agnostic accuracies.

Significantly, our findings highlight an apparent trade-off between known and unknown accuracies. By optimally tuning these parameters, especially with an $\alpha\approx1$ when $\beta=2$ and $\beta=5$, we can enhance known accuracy without adversely impacting the unknown accuracies.

\section{Conclusion}
\vspace{-2.5mm}

In this work, we turn our attention to the Long-tailed Generalized Category Discovery (Long-tailed GCD), where the unlabeled data predominantly consists of known classes, leading to an inherent imbalance. To counteract this challenge, we proposed a simple and effective method fortified by two regularizations: tail reweighting and a class prior constraint. Through comprehensive experimentation covering both balanced and multiple imbalanced contexts, our method emerges as a formidable contender. It consistently matches or exceeds the performance of leading-edge solutions across a variety of imbalanced settings, underscoring its adaptability and paramount effectiveness in real-world applications.

%

{\small
\bibliographystyle{ieee_fullname}
\bibliography{ncd}
}

\end{document}